\documentclass[runningheads]{llncs}
\usepackage{graphicx}
\usepackage{todonotes}
\usepackage{xspace}
\makeatletter
\DeclareRobustCommand\bmvaOneDot{\futurelet\@let@token\bmv@onedotaux}
\def\bmv@onedotaux{\ifx\@let@token.\else.\null\fi\xspace}
\makeatother
\def\eg{\emph{e.g}\bmvaOneDot}

\def\ie{\emph{ie}\bmvaOneDot}

\def\wrt{w.r.t\bmvaOneDot}

\def\vs{\emph{vs}\bmvaOneDot}

\begin{document}
\title{A Token-wise CNN-based Method For Sentence~Compression}
\author{Weiwei Hou\inst{1}\textsuperscript{,}\thanks{This work is supported by the Australian National University. This paper is published at ICONIP 2020.} \orcidID{0000-0002-0758-2870} \and
Hanna Suominen\inst{1,2,3} \orcidID{0000-0002-4195-1641} \and Piotr Koniusz\inst{1,2} \orcidID{0000-0002-6340-5289
}\and Sabrina Caldwell\inst{1} \orcidID{000-0003-0605-3149} \and Tom Gedeon\inst{1} \orcidID{0000-0001-8356-4909}} 
\authorrunning{Hou et al.}
\institute{The Australian National University \\ \email{\{weiwei.hou, hanna.suominen, sabrina.caldwell, tom.gedeon\}@anu.edu.au}\and Data61/CSIRO, Australia \\
\email{piotr.koniusz@data61.csiro.au} \and
University of Turku Finland} 
\maketitle              % typeset the header of the contribution
\begin{abstract}
Sentence compression is a \textit{Natural Language Processing} (NLP) task aimed at shortening  original sentences and preserving their key information. Its applications can benefit many fields \eg, one can build tools for language education. However, current methods  are largely based on \textit{Recurrent Neural Network} (RNN) models which suffer from poor processing speed. To address this issue, in this paper, we propose a token-wise\textit{ Convolutional Neural Network},  a CNN-based model along with pre-trained 
\textit{Bidirectional Encoder Representations from Transformers}
(BERT) features for deletion-based sentence compression. We also compare our model with RNN-based models and fine-tuned BERT. Although one of the RNN-based models outperforms marginally   other models given the same input, our CNN-based model was ten times faster than the RNN-based approach.

\end{abstract}
\section{Introduction}
Deletion-based sentence compression refers to the task of extracting key information from a sentence by deleting some of its words. It is often used as an initial step for generating document summaries or as an intermediary step in information retrieval and machine translation. Its applications may also benefit many fields such as e-Learning and language education. 

Recent studies focus on adapting  methods based on neural networks to solve deletion-based sentence compression as a sequential binary classification problem  (see Section~\ref{RW}). \textit{Recurrent Neural Networks} (RNNs) are one of the most popular network architectures that handle sequential tasks. Models such as \textit{Gated Recurrent Units} (GRUs), \textit{Long-Short Term Memory} (LSTM) networks, and BiLSTM \textit{Bidirectional LSTMs} (BiLSTMs) are found to be suited for this task. However,  training these RNN-based models can be time consuming. Applications with poor response speed are the cause of negative user experience. 

In contrast, \textit{Convolutional Neural Networks }(CNNs) outperform RNNs in their training speed and reportedly have a similar or better performance than RNN-based models in many tasks \cite{Wenpeng2017,Shaoji2018}. They are widely applied to tasks such as object recognition in Computer Vision. In \textit{Natural Language Processing} (NLP), CNNs have been studied for document summarization, sentiment analysis, and sentence classification, among others. However, the majority of these methods concern the sentence or document level.

In this paper, we apply CNNs to sentence compression at the token~level. However, CNNs are weaker at capturing sequential information. To circumvent this issue, we train our model with pre-trained \textit{Bidirectional Encoder Representations from Transformers} (BERT) \cite{BERT2019} features. In addition, we also compare our model performance against RNN and BERT fine-tuned models. We test the performance in both the correctness and efficiency.  

\section{Related Work}\label{RW}

\subsection{Recurrent Neural Networks}

RNN-based approaches are widely applied to sequential problems such as machine translation, sentence parsing, and image captioning. Inspired by these core NLP tasks, \cite{Filippova2015} concatenated input sequence and its labels by a key word `GO' as an input to a sequence-to-sequence framework. The goal was to predict sequence labels for each word succeeding `GO'. Their network architecture is composed of three layers of LSTMs. On top of the LSTM layers is a softmax layer that produces  final outputs. Their method takes on input only word embeddings in the form of the vector representations. 

Subsequently, authors of \cite{Liangguo2017} and \cite{Yang2017} discovered that adding syntactic information improves the performance. Both works included the \textit{Part-of-Speech} (POS) tags, dependency type, and word embeddings. The results showed significant accuracy improvements. In addition, instead of concatenating embedding sequences and labels, both these studies %works directly 
used hidden vectors to predict labels. The difference between these two methods is that approach \cite{Liangguo2017} uses a framework with  three layers of BiLSTMs while approach \cite{Yang2017}
has a more complex architecture which includes a layer of 
\textit{Bi-directional RNNs} (BiRNNs) and a layer of Graph Neural Networks. A year later, authors of \cite{Yang2018} proposed an approach based on reinforcement learning which includes a BiRNN-based policy network containing two types of actions -- REMOVE and RETAIN, where words marked as REMOVE are  deleted to obtain a sequence of predicted compression. Such a sequence is then fed into a syntax-based evaluation step  that  examines the predicted compression according to two rewards. The first reward concerns the fluency of generated sentence against a syntax-based language model. The second reward is based on the comparison of the generated sentence compression rate with an average compression rate. Their method works well on both large unlabeled and labeled datasets. However, such a model is difficult to train. 

\subsection{Bidirectional Encoder Representations from Transformers}
BERT is a language representation model which takes the word position into account to capture a word with its context \cite{BERT2019}. Unlike 
\textit{Global Vectors for Word Representation} (GloVe), Word2Vec, and many other 
context independent word embeddings, BERT not only embeds semantic information but it also captures the structural information in a sequence. In addition, BERT enables bi-directional prediction. It uses a \textit{Masked Language Model} mechanism that randomly masks a certain percentage of tokens in an input sequence. The objective is to use both the preceding and succeeding context to predict the masked token. 

Apart from providing pre-trained language representations, BERT can  be fine-tuned on related tasks. It has reportedly achieved the state-of-the-art performance in nine NLP tasks~\cite{BERT2019}. BERT is said to have also the ability to capture high-level linguistic knowledge (\eg, semantics, syntax, or grammar)  \cite{Clark:2019,Goldberg:2019,Liu:2019}. 

To the best of our knowledge, our is the first work using BERT for deletion-based sentence compression tasks. We compare pre-trained BERT layers with word embeddings, POS embeddings, and dependency embeddings given the same network architecture to explore whether BERT is able to capture complex syntactic information.

\section{Method}
We define the deletion-based sentence compression task as a token level segmentation task. Specifically, we have given a sequence of \(s = \{w_1, \, w_2, \, w_3, \, \cdots \, , \, w_i\}\) as an original sentence, where  \(i\) is the number of tokens in this sequence, for \(s\), and we have a corresponding sequence of the mask \(y=\{y_1, \, y_2, \,y_3, \, \cdots \, , \, y_i\}\), where $y_i\in \{ 0, 1 \}$ is the ground truth label of \(w_i\). Moreover, by zero (one) we mean that a token needs to be deleted (retained) from the original sequence. The goal is to train a model to predict whether \(w_i\) in sequence \(s\) should be deleted or~retained.

\begin{figure}[t]
    \centering
    \includegraphics[scale=0.4]{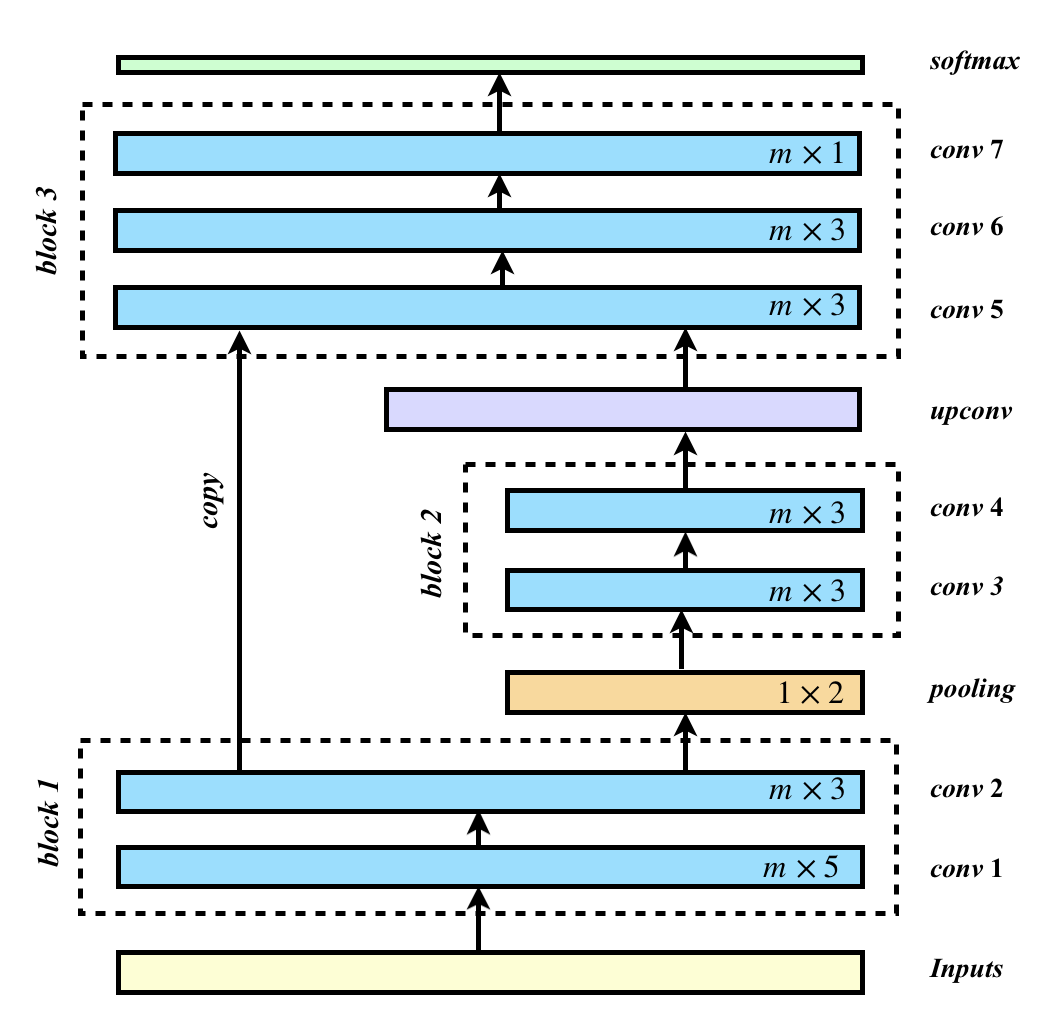}
    \caption{Graphical illustration of a CNN-based network architecture.  Convolutional layers are denoted by \textit{conv} 1, \textit{conv} 2, \textit{conv} 3, $\ldots \, , $\textit{conv} 7.}
    \label{fig:my_label}
\end{figure}

\subsection{Network Architectures}
\label{our_model}
Our approach is largely based on U-Net~\cite{Ronneberger2015}, with some differences. The network was originally designed for pixel-level image segmentation tasks. It is a fast and lightweight network which shows extraordinary performance in many image segmentation tasks. The reason we choose U-Net as our base network is that our task is a token level binary classification. Regular CNN cannot capture various levels of coarseness of information and ``expand" it back to the token level for segmentation. In addition, we believe that the max-pooling operation can be seen as realizing ``compression". 

We adapted the original U-Net network architecture to train a model for a text-based task. As Fig.~\ref{fig:my_label} shows, we assume we have   word \(w_i\) in sequence \(s\) where each \(w_i\) has associated with it a $j$-dimensional vector \(w_i=\{e_{i,1},\, e_{i,2}, \, e_{i,3}, \, \cdots \, , e_{i,j}\}\). This setting yields a matrix  of size \(i \times j\) which forms the input to the first layer of our network. 

Fig.~\ref{fig:my_label} shows (from the bottom to the top) that the network architecture contains three main blocks. The first block has two standard convolution layers (conv 1 and conv 2). The conv 1 layer has a \(m\times5\) kernel; $m$ is determined by the number of input channels (see Section \ref{experiment}) and conv 2 has a  \(m\times3\). 

Moreover, between the first and second block, there is a \(1\times2\) pooling layer (pool1) with the stride size of 2 for down-sampling. The pooling output is fed to the second block which contains two \(m \times 3\) convolution layers (conv3, conv4). After the above operations, the output is halved in size. 

In order to map the output into a token level, we up-sample (deconv) the output from the second block to double back the output size and then the output is concatenated with the previous output from the first block. This operation allows the network to remember ``early''  features.  The third block includes two regular $m\times3$ convolutions (conv5, conv6) and one $m\times1$ convolutions (conv7). All above convolutions are followed by a \textit{Rectified Linear Unit} (ReLU). 

Lastly, we pass the convolution output \(x=\{x_1, \, x_2, \, x_3, \, \cdots \, , \, x_i\}\) though a token-wise softmax layer to predict the mask \(y_i\) as follows:
\begin{equation}\label{eq1}
P(y_i|x_i)=softmax(Wx_i),
\end{equation}
where $W$ is a weight matrix to be learned. 

We introduce two main modifications to the original network architecture \cite{Ronneberger2015}. The first one is that instead of using the same kernel sizes, our network uses a mix of kernel sizes in order to  capture multi-gram information. The second change is that we reduced the depth of the network due to the size of our sentences being much smaller than image sizes from the original work. Hence, empirically, a shallower network tends to perform better.

\section{Experiments}
\label{experiment}
\subsection{Data}
For our experiments, we use for the GoogleNews dataset, an automatically collected dataset according to approach  \cite{Filippova2013}. The currently released data contains $200,000$ sentence pairs, written in ten files. One file includes $10,000$ pairs for evaluation while remaining $190,000$ pairs are contained in other files.

\vspace{0.05cm}
\noindent{\textbf{\textit{GoogleNewsSmall}}}. For parts of our experiments, we use only the $10,000$ pairs based evaluation set. We call it \textit{GoogleNewsSmall}. The reason that we choose \textit{GoogleNewsSmall} is because one of purposes of this study is to compare the performance of CNN- and RNN-based models on deletion-based sentence compression tasks. Therefore, to ensure fairness, \textit{GoogleNewsSmall } includes exactly the same dataset as the previous method settings \cite{Liangguo2017}. Furthermore, we use the first $1,000$ sentences as the testing set, the next $1,000$ sentences as the validation set, and the remainder as the training set.

\vspace{0.05cm}
\noindent{\textbf{\textit{GoogleNewsLarge}}}.  We are also interested in discovering the impact of the training data size on performance. Therefore, in the second setting, we include the entire $200,000$ sentence pairs in our experiments. We denote this setting as \textit{GoogleNewsLarge}. For testing and validation, we use the  setting already described above. The remaining $198,000$ pairs are used as the training set.

\subsection{Experimental Setup} 

\noindent{\textbf{\textit{BiLSTM+Emb}}}.
\label{bilstm}
In this experiment, we use the same network settings as the one described in \cite{Liangguo2017}, which includes a base model architecture with three BiLSTM layers and a dropout layer between them. The input to this network is contains word embeddings only. The purpose of this experiment is to train a base-line RNN-based network.

\vspace{0.05cm}
\noindent{\textbf{\textit{CNN+Emb}}}.
\label{base-CNN}
In this experiment, we use our method proposed in Section \ref{our_model}. For input, we again use word embeddings as the only input to ensure the comparison with RNNs in the same testbed.

\vspace{0.05cm}
\noindent{\textbf{\textit{BiLSTM+BERT}}}.
\label{bilstm-bert}
In this experiment, we adapt the setting of
\textbf{BiLSTM+}\newline\textbf{Emb} by replacing the word embeddings  with the last layer of pre-trained BERT outputs. We also record the training time, in seconds, from the point of starting training to full convergence. 

\vspace{0.05cm}
\noindent{\textbf{\textit{CNN+Multilayer-BERT}}}.
\label{cnn_multi}
This experiment is composed of a group of sub-experiments. We use the same network architecture as in Section \ref{base-CNN}. We firstly extract the last four BERT layers (-1, -2, -3, -4) from a pre-trained model. Then, we run the experiments by feeding the network with four different input layer settings:
(i) input with the layer -1, 
(ii) input with the layer -1, and -2, 
(iii) input with the layer -1, -2, and -3), and
(iv) input with all last four layers. %does not improve the performance. 
We record the training time of experiment with the layer -1 as input, from the point of start of training to the full converge. 

\subsection{Experiments on Different Network Settings}
This experiment aims to test the model performance \wrt reducing network layers. In the first sub-experiment, we remove one convolution layer from each block (conv 2, 4, and 5 layers from the original settings) in order to test the impact of convolution layers. In the second sub-experiment, we remove pooling, conv 4, 5 of block 2, and upsampling layers. The purpose of this experiment is to compare the performance of a stack of convolution layers and our model.

\subsection{Experiments \wrt the Training Size}
This experiment evaluates the performance \wrt different sizes of training data. We vary the training size between $8,000$ and $19,800$. We divide the training progress into ten steps. At each step, the training size increases by $20,000$. We detail this experiment in Section \ref{cnn_multi} by only testing on the ``last four layers'' setting.

The data is labeled as described in approach \cite{Filippova2013}. Word embeddings are initialised by GloVe 100-dimensional pre-trained embeddings \cite{jeffery2014}. BERT representations are extracted from the 12/768-uncased-base pre-trained model\footnote{https://github.com/google-research/bert}.
All experiments are conducted on a single GPU (Nvidia GeForce GTX Titan XP 12GB). For CNN-based networks, all input sequences have a fixed length of 64\footnote{We concluded that it is the best input size after calculating the mean and max length of all sentences in the entire dataset, and evaluating the efficiency of extracting BERT features.}.
The number of input channels is and layers are the same as in BERT.

\subsection{Quantitative Evaluations}
%\subsubsection{Automatic Evaluation.}

To evaluate the performance, we use the accuracy and F1 measure.
%, which are 0 for deleted words and 1 for retained words. 
F1 scores are derived from precision and recall values \cite{Chinchor1992}
where the precision is defined as the percentage of predicted labels (1 = retained words) that match ground truth labels, and recall is defined as the percentage of labeled retained words in ground truth that overlap with the predictions.

Regarding the training efficiency, we evaluate it as follows. Firstly,  we compare both the F1 scores and accuracy of CNN and RNN-based model as a function of each recorded time point. Secondly, %to achieve the same accuracy, 
we compare the training time of two models.  

\subsection{Perception-based Evaluations}
In order to test the readability of  outputs and their relevance to the original inputs, we asked two native English speakers to manually evaluate outputs from the \textbf{CNN-Multilayer-BERT} and \textbf{BiLSTM-Emb} (baseline). The inputs are the top 200 sentences from the test set. Evaluation methods follow approach~\cite{Filippova2015}. 
 \begin{table}
\begin{minipage}{0.5\textwidth}
         
\caption{Results given different\newline experimental settings for models\newline trained on \textit{Google-} \textit{NewsSmall}.}\label{result-table}
\begin{tabular}{|l|c|}
\hline
\textbf{Method } & \textbf{F1} \\ \hline
BiLSTM+Emb & 0.74 \\
BiLSTM+BERT & 0.79 \\
BiLSTM+SynFeat\cite{Liangguo2017} & 0.80\\
LK-GNN+Emb\cite{Yang2017} & 0.80 \\
LK-GNN+All Features\cite{Yang2017} &  \textbf{0.82} \\
\hline
CNN+Emb & 0.72\\
CNN+Multilayer-BERT & 0.80\\
\hline
\end{tabular}
\end{minipage}
\begin{minipage}{0.4\textwidth}
\caption{\label{bert_layers}  Results \wrt using different layers of BERT features (trained on \textit{GoogleNewsSmall}).}

\begin{tabular}{|c|c|c|}
\hline \textbf{BERT layers} &  \textbf{F1}  & \textbf{Accuracy} \\ \hline
            Only -1 & 0.78 & 0.80 \\
            -1, -2 & 0.79 & 0.81 \\
            -1, -2, -3 & 0.80 & 0.81 \\
            -1, -2, -3, -4 & \textbf{0.80} & \textbf{0.82} \\
            \hline
            \end{tabular}

        \end{minipage}
    \end{table}

\begin{table}
\caption{\label{conv_layers}  Results given different network settings (trained on \textit{GoogleNewsSmall}).} 
\centering
\begin{tabular}{|c|c|}
\hline \textbf{Network Setup} &  \textbf{F1}  \\ \hline
            remove conv 2, 4, and 5 &  0.797  \\
            remove pooling, conv 4,5 and upsampling
             &  0.786 \\
             original setting (channel size conv4=128, conv6=256)&0.805\\
            \hline
            \end{tabular}

\end{table}
\section{Results}\label{results}
We report the accuracy of different experimental settings in Table \ref{result-table}.
We note that \textbf{LK-GNN+All Features} achieves the best resukts.  \textbf{CNN+Multilayer-BERT} performs the same as \textbf{BiLSTM+SynFeat} and \textbf{LK-GNN+Emb} ($0.80 = 80\%$). Next, we compare the performance of different models with the same input settings. When model inputs are word embeddings, the table shows that the RNN-based model outperforms the CNN-based one. The results reflect our assumption that CNNs are weaker in capturing sequential information compared to RNNs. However, the performance gap between \textbf{CNN+Emb} and the \textbf{BiLSTM+Emb} is not significant. Therefore, CNN-based models are a reasonable choice. 

Looking at experiments that use the same network architecture but different input settings, we notice that models with BERT used as the input have significantly better performance than the models with Emb inputs. When comparing BERT to add-on syntactic features,  \textbf{BiLSTM+BERT} slightly under-performs \textbf{BiLSTM+SynFeat} (\ie, $1\%$ lower on F1). This implies that BERT captures both syntactic and semantic information. Moreover, when we test multiple layers of BERT features, we can see that \textbf{CNN+Multilayer-BERT} performs the same as the model with add-on syntactic features. Therefore, multiple layer BERT features enhance  learning ability of the model.  
\begin{table}[]
    \centering
     \caption{\label{manual-evl}  Perception-based Evaluations (trained on \textit{GoogleNewsSmall}).}
    \begin{tabular}{|l|c|c|}
        
\hline
\textbf{Method } & \textbf{Readability}&\textbf{Informativeness} \\ \hline
BiLSTM+Emb & 4.0 & 3.4\\
CNN+Multilayer-BERT & 4.3 &3.6\\
\hline
    \end{tabular}
\end{table}
\begin{table}
\centering
\caption{\label{runtime-table} Training time of CNN-  \vs LSTM-based models. The first layer of BERT output is used as input to the model.}
\begin{tabular}{|c|cc|cc|}
\hline \textbf{Run time(s)}& \multicolumn{2}{c|}{\textbf{CNN+BERT}} & \multicolumn{2}{c|}{\textbf{LSTM+BERT}} \\ 
&F1& Accuracy  &F1& Accuracy \\
\hline
16&0.60 & 0.73 & 0.60 & 0.66\\
64&0.75 & 0.73 & 0.50 & 0.64 \\
120&0.766 & 0.79 & 0.59 & 0.67 \\
210&0.78 & 0.80 & 0.50 & 0.69\\
720&- & - & 0.70&0.76\\
1095&- & - & 0.73&0.78\\
1483&- & - & 0.75&0.78\\
1863&- & - & 0.75&0.79\\
2239&- & - & 0.78&0.80\\
2622&-& - & 0.79&0.79\\
3303&-& - & 0.80&0.81\\
\hline
\end{tabular}
\end{table}
To further investigate BERT, Table \ref{bert_layers} shows the impact of using different layers of BERT features on the model performance. The model trained with last four BERT output layers achieves best results (F1 = 0.80, Accuracy = 0.82). However, the model that only uses the last layer of BERT under-performs by $2\%$ (F1 score) and by $1\%$ accuracy. We notice a trend that increasing the number of BERT layers will result in a slightly better performance. This supports our hypothesis that using multiple layers of BERT features  improves the performance. Note that we tested multiple layers of BERT features on our CNN based model but not on RNN-based models because concatenating features of four BERT output layers results in an extremely long input (each layer of BERT has a hidden size equal \(4\times768\)). Thus, the computation cost would be very expensive. %Finally, we also conducted experiments to test the training efficiency of both types of network. 
Moreover, Table \ref{manual-evl} shows results for the perception-based evaluations of our model compared to the baseline. As one can see, the model trained with BERT scores better on readability and informativeness.

\begin{figure}
    \centering
    \includegraphics[scale=0.5]{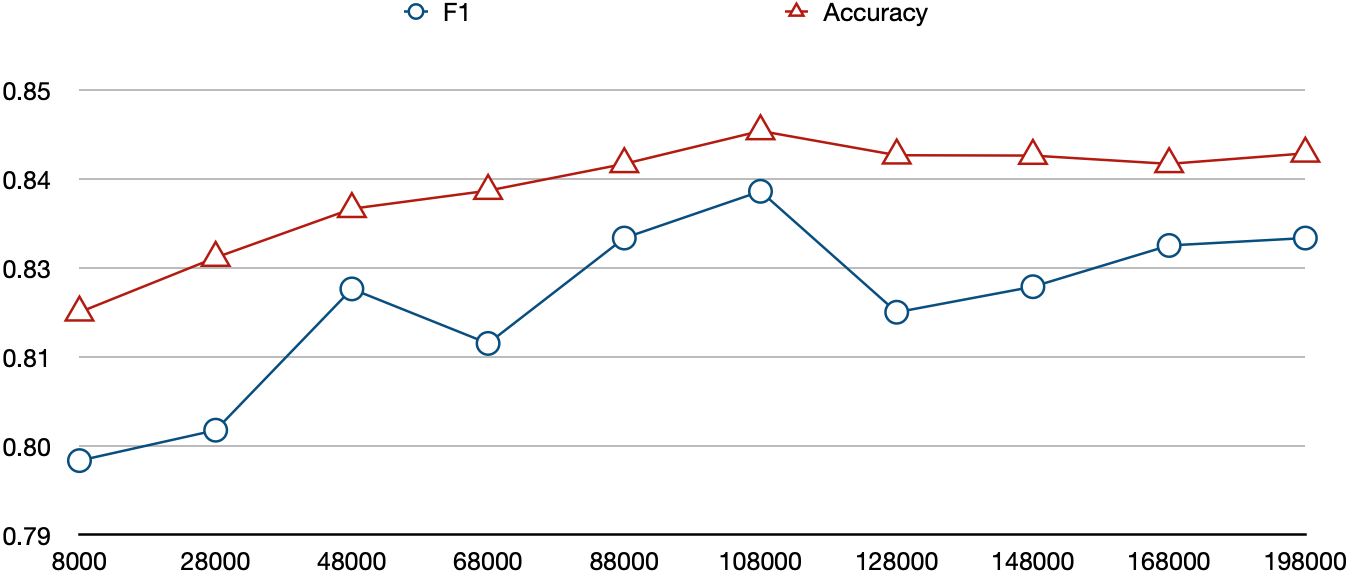}
    \caption{Performance of CNN+Multilayer-BERT (last four layers used) in terms of F1 and accuracy. Experiments are conducted on \textit{GoogleNewsLarge}.}
    \label{fig2}
\end{figure}

 The comparisons are demonstrated in Table \ref{runtime-table}. As we expected, the CNN-based model converged  faster when compared to  the BiLSTM-based model. It reached its best accuracy performance (F1 = 0.78, Accuracy = 0.81) after 210 seconds. However, while reaching the same F1 score, the BiLSTM-based model needs 10 times longer training time. To outperform the CNN-based models in terms of both the F1 and accuracy, the BiLSTM model took 15 times more time to train. Therefore, our model significantly outperforms BiLSTM models if F1, accuracy and time are taken into account. Table \ref{conv_layers} shows the F1 scores for different network settings. As we can see, by removing three convolutional layers, the model slightly under-performs the original model. However, when the pooling, conv 4, conv 5 and upsampling were removed, the results are nearly 2\% lower compared to the original settings. This result supports our assumption that pooling help extract compressed information. 

In addition, we also investigated the impact of the training data size on the CNN-based model. Figure \ref{fig2} shows the F1 scores and accuracy of our CNN-BERT model (\textbf{CNN+Multilayer-BERT}) \wrt different size of training data.  We observed that by increasing the training size to approximately $100,000$ pairs, the model performs reaches over 0.84 in both the F1 score and accuracy. For more training data, the results show no further improvement and when we increase the training size to be equal to the size of full training set, the results  drop to 0.83 (F1 score). We believe that this is caused by the noise in the dataset as authors of approach \cite{Yang2018} note that this dataset is automatically labeled based on syntactic-tree-pruning method. Noise can be introduced by syntactic tree parsing errors. Approach \cite{Yang2018} scores 0.84 (F1 score) on their LSTM models with the same training data as in our experiments. However, under the same settings, a model trained on 2 million training pairs reported 0.82 (F1 score) \cite{Filippova2015}. Authors of approach \cite{Yang2018} also augured that this training data may contain errors  caused by syntactic tree parsing errors during data annotation. However, we do not evaluate the quality of the data. Our results indicate that the effective training size equals $100,000$.

\section{Discussion}
Section \ref{results} showed that, the RNN-based models slightly outperform CNN-based models in their correctness; on average, their F1 scores were 2\% higher. However, the CNN-based models performed over ten times faster. In addition, to improve the results of CNN-based models, we adopted BERT features as our networks inputs. The results showed that the model with four layers of BERT features achieved equal performance compared to approach with add-on syntactic features \cite{Liangguo2017} given training size equal $8,000$. It implies that multiple layers of BERT capture both the syntactic and semantic information. 
We argue that the CNN with multiple layers of BERT features was quite a reasonable setting. Since each layer of BERT features has a vector size of 678, concatenating multiple BERT layers as inputs for RNN-based models is computationally prohibitive.

In addition, we also tested the impact of the training data size on our \textbf{CNN}\\\textbf{+Multilayer-BERT} model given four distinct  feature settings of BERT, and we found that the F1 scores do not improve further when the training date size reaches approximately $100,000$ pairs. We believe that such a result is caused by the noise in the dataset. We believe that such a noise was introduced during the data collecting and labeling process. Similar observations were reported by authors of approaches \cite{Yang2017,Yang2018}. In contrast to previous works, we report what is a reasonable trade-off in terms of the dataset size. 

In this paper, we did not directly compare our results with the  state-of-the-art model \cite{Yang2018} for two main reasons. Firstly, our work mainly focused on comparing the performance on different base model settings. Authors of approach \cite{Yang2018} proposed a reinforcement learning method, implementing bidirectional RNNs as the base model of the policy network, and this setting is quite similar to approach \cite{Liangguo2017}. Secondly, we tested methods that do not include any domain specific knowledge. One of reward rules in their method uses scoring the sentence compression rates.  Since the data was generated by predefined rules, adding such a rule could improve the performance. Although we did not directly compared our results with theirs, we reported results given the  training size of $100,000$. Our method reaches 84\% (F1 score) which is equal to their implemented LSTM model, and only about 1\% lower than their reinforcement-based method. We believe if the base model was replaced with our CNN model, the final accuracy would be similar while enjoying faster training. %This would be a useful experiment in future work. 

\section{Conclusions}
In this paper, we studied the performance of CNN-based models for the deletion-based sentence compression task. We first tested the  correctness results against the most commonly implemented RNN-based models as well as fine-tuned BERT. Subsequently, we examined the training efficiency of both models. We also compared the results when using a pre-trained BERT language representation as an input to the models with classical word-embeddings and/or other add-on syntactic~information. 

Our results show that the CNN-based model requires much less training time than the RNN-based model. In addition, the pre-trained BERT language representation 
highlighted its ability to capture deeper information compared to classical word embedding models. BERT could also serve as a replacement of manually introduced add-on syntactic information.  Finally, we observed that increasing the size of training data beyond certain point does not improve the performance further. %The reasonable training size of training data was about half of its size.

Our approach can potentially reduce the cost of building sentence compression applications such as language education tools. Our approach saves computational resources which promotes interactive applications while preserving their accuracy. In the future, we will use our model as a backbone in a reading assistant tool supporting university \textit{English as a Second Language } (ESL) students in their reading activities. We will also continue to study sentence compression with the focus on approaches that can customise the output.


\end{document}